\documentclass[letterpaper, 10 pt, conference]{ieeeconf}  % Comment this line out if you need a4paper

\IEEEoverridecommandlockouts                              % This command is only needed if 
\usepackage{amsfonts}
\usepackage{mathtools}
\usepackage[noadjust]{cite}
\usepackage{booktabs,multirow}
\usepackage{hyperref}

\title{\LARGE \bf
An Attentional Recurrent Neural Network for Occlusion-Aware Proactive Anomaly Detection in Field Robot Navigation
}

\author{Andre Schreiber, Tianchen Ji, D. Livingston McPherson, and Katherine Driggs-Campbell% stops a space
\thanks{This work was supported in part by the National Robotics Initiative 2.0 (NIFA\#2021-67021-33449). The robot platforms and data were provided by the Illinois Autonomous Farm and the Illinois Center for Digital Agriculture.}
\thanks{The authors are with the Coordinated Science Laboratory, University of Illinois at Urbana-Champaign, Champaign, IL 61820 USA (e-mail: \{andrems2,tj12,dlivm,krdc\}@illinois.edu). \textit{(Corresponding author: Andre Schreiber.)}}
}

\begin{document}

% To fix weird citation
\bstctlcite{IEEEexample:BSTcontrol}

\maketitle
\thispagestyle{empty}
\pagestyle{empty}

%%%%%%%%%%%%%%%%%%%%%%%%%%%%%%%%%%%%%%%%%%%%%%%%%%%%%%%%%%%%%%%%%%%%%%%%%%%%%%%%
\begin{abstract}
The use of mobile robots in unstructured environments like the agricultural field is becoming increasingly common. The ability for such field robots to proactively identify and avoid failures is thus crucial for ensuring efficiency and avoiding damage. However, the cluttered field environment introduces various sources of noise (such as sensor occlusions) that make proactive anomaly detection difficult. Existing approaches can show poor performance in sensor occlusion scenarios as they typically do not explicitly model occlusions and only leverage current sensory inputs. In this work, we present an attention-based recurrent neural network architecture for proactive anomaly detection that fuses current sensory inputs and planned control actions with a latent representation of prior robot state. We enhance our model with an explicitly-learned model of sensor occlusion that is used to modulate the use of our latent representation of prior robot state. Our method shows improved anomaly detection performance and enables mobile field robots to display increased resilience to predicting false positives regarding navigation failure during periods of sensor occlusion, particularly in cases where all sensors are briefly occluded. Our code is available at: \url{https://github.com/andreschreiber/roar}.
\end{abstract}

%%%%%%%%%%%%%%%%%%%%%%%%%%%%%%%%%%%%%%%%%%%%%%%%%%%%%%%%%%%%%%%%%%%%%%%%%%%%%%%%
\section{INTRODUCTION}\label{sec:intro}

Throughout various domains, mobile robots are becoming increasingly prevalent as technological advancements enable such robots to autonomously execute a greater number of tasks. In agriculture, for example, compact mobile robots can move between crop rows and have been used to perform tasks such as corn stand counting~\cite{embedded} and plant phenotyping~\cite{phenotype}. However, the agricultural field environment presents numerous challenges for such robots as this unstructured environment displays cluttered foliage, varying lighting conditions, and uneven terrain.

\begin{figure}[tp]
  \centering
  \includegraphics[width=3.4in]{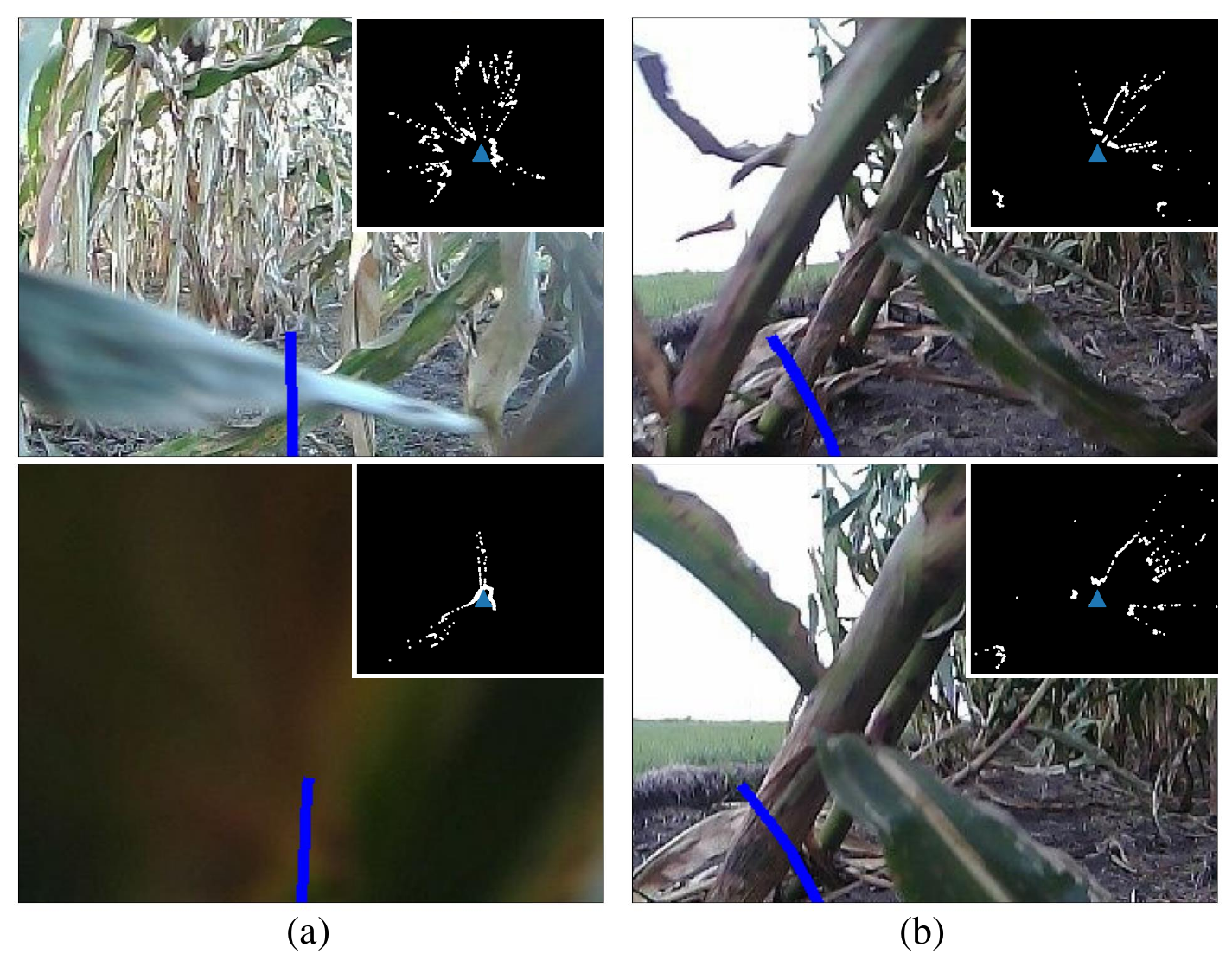}
  \caption{Example sequences from the field environment dataset introduced by Ji \textit{et al.} \cite{paad}, with each of the two sequences displayed top to bottom. A blue line indicates the planned trajectory, and the LiDAR map is shown in the top right of the images. (a) shows a brief occlusion caused by low-hanging vegetation which does not lead to immediate navigation failure, while (b) shows an obstruction that leads to navigation failure.}
  \label{failvsocc}
\end{figure}

These challenging conditions require algorithms that are robust to noise and sensor occlusions in order for the robots to remain autonomous, especially as the difficult nature of the environment increases the possibility that robots enter failure modes. Entering such failure modes may lead the robot to require external intervention to accomplish its task or may involve damage to the robot~\cite{tsp}. Thus, detecting potential navigation failures ahead of time becomes increasingly important in order to prevent damage and ensure optimal efficiency. However, the difficulty of developing algorithms to proactively detect such failure modes is exacerbated by the unstructured nature of the field environment, as such algorithms must be able to differentiate between scenarios representing genuine navigation failures (e.g., colliding with rigid obstacles or prematurely leaving the crop row) and the frequent but ultimately non-catastrophic noise (e.g., occlusions) created by the environment. An example of the difference between an occlusion that does not lead to navigation failure and a true failure mode is shown in Fig \ref{failvsocc}.

Detecting such failure modes is commonly approached from the perspective of anomaly detection (AD) \cite{encdecad, vaelstm, paad, svae, graspe}, with failure modes being treated as anomalies. Many works on AD \cite{encdecad, svae, vaelstm} view the problem from a reactive perspective, in which anomalies are detected as they occur; however, with reactive AD, potential failure conditions cannot be detected before they occur in order to avoid them. Due to this limitation, recent work has focused on proactive AD in which the robot predicts the probability of failure within a time horizon using both current sensory inputs and planned future actions \cite{land, badgr, graspe, paad}. In unstructured environments like those seen by the field robots, AD models may additionally utilize multi-sensor fusion of different inputs like RGB cameras and LiDAR to increase robustness to noise and occlusions~\cite{graspe, paad}. In these multi-sensor approaches, occlusion conditions can be implicitly learned without supervision \cite{paad} or explicitly modeled \cite{graspe}. However, such approaches only make use of current sensory inputs and do not maintain a history of prior sensory state. Thus, using multi-sensor fusion to address the possible occlusion of sensors can still fail if all sensors are briefly occluded. Such failures during periods of occlusion typically manifest as false positives, where the AD algorithm falsely predicts that the robot has entered a failure mode. As a prediction of failure can lead to interruption of normal robot operation, reducing these false positives minimizes the number of spurious interruptions and improves operating efficiency.

Due to these limitations, we introduce a new proactive AD neural network. Our proposed network, termed Recurrent Occlusion-AwaRe (ROAR) anomaly detection, learns when sensors are occluded and incorporates an attention-based mechanism to fuse sensor data, predicted future actions, and a summary of prior robot state (informed by occlusion of the sensors) to provide improved AD performance when sensor occlusion occurs. Our architecture can reduce false positives in brief periods of total sensor occlusion, and more effectively makes use of its summarization of the prior robot state by explicitly learning when sensors are occluded. Furthermore, as multi-sensor approaches may not always be possible or economical due to their increased hardware requirements, we demonstrate that a variant of our model using only one input sensor can still provide an attractive alternative to existing multi-sensor fusion approaches.

We summarize our contributions as follows:
\begin{enumerate}
    \item We propose an attention-based recurrent neural network architecture that fuses planned future actions and multiple sensory inputs with a latent representation of robot state (which summarizes prior sensory inputs and prior planned control actions) to improve AD performance in unstructured environments, particularly when total sensor occlusion occurs.
    \item We leverage an explicitly learned model of sensor occlusions to provide enhanced utilization of the latent representation of robot state and improved AD performance with our recurrent neural network architecture.
    \item We show that our network demonstrates improved performance over existing methods, and displays increased robustness against false positives in brief periods of total sensor occlusion. We also demonstrate that when even only one sensor modality is used, our model significantly outperforms other single-sensor networks and provides an attractive alternative to multi-sensor fusion models in cases where multiple sensors may not be available.
\end{enumerate}

\section{RELATED WORK}\label{sec:related}
AD is studied and applied in a variety of contexts. In the context of robotics and autonomous systems, AD (also called outlier detection or novelty detection) is frequently used to detect failures and often draws upon additional areas such as deep learning and multi-sensor fusion to provide improved performance.

\subsection{Multi-Sensor Fusion using Neural Networks}
Contemporary robots and autonomous systems typically feature numerous sensors. Thus, there has been pronounced study on how to effectively fuse such sensor data, with many approaches utilizing neural networks for such fusion. For example, Nguyen \textit{et al.} \cite{nmfnet} propose a neural network architecture that fuses multi-modal signals from images, LiDAR, and a laser distance map in order to learn to navigate in complex environments like collapsed cities. Similarly, Liu \textit{et al.} \cite{learnetoe} present a method for learning navigation policies that makes use of multi-modal sensor inputs, improving robustness to sensor failure by introducing an auxiliary loss to reduce variance of multi- and uni-sensor policies and by introducing sensor dropout. Likewise, Neverova \textit{et al.} \cite{moddrop} present ModDrop, which introduces a modality-wise dropout mechanism similar to sensor dropout for multi-modal gesture recognition. ModDrop randomly drops sensor modality components during training, improving model prediction stability when inputs are corrupted.

Numerous deep learning-based methods of multi-sensor fusion integrate attention mechanisms to achieve more effective fusion of sensory inputs. Such attention-based fusion mechanisms have seen significant use in fields such as human activity recognition. For example, \cite{attnsense, actrecimu, marfusion} all describe attentional fusion architectures that combine data collected from multiple sensors that are affixed to a subject's body.

\subsection{Occlusion Modeling}
In robotics and autonomous systems, sensors can experience occlusions, which may require a fusion mechanism with special provisions to ensure stable predictions. Occlusions may manifest as faulty sensor readings, and sensor- or modality-wise dropout could be used to account for such occlusions \cite{learnetoe, moddrop}. Other works have also attempted to specifically devise strategies for fusion under sensor occlusion rather than treating an occlusion as a sensor failure. For example, Palffy \textit{et al.} \cite{occped} introduce an occlusion-aware fusion mechanism for pedestrian detection by using an occlusion-aware Bayesian filter. Ryu \textit{et al.} \cite{confocc} describe a method for robot navigation that is intended to work specifically in cases of prolonged sensor occlusion of 2D LiDAR caused by issues like dust or smudges. However, the assumption of prolonged occlusions does not entirely suit the agricultural field environment that we target, which typically features brief dynamic occlusions (e.g., a leaf briefly covering the camera as the robot drives down a crop row). Similarly, we seek to design a multi-modal anomaly detector, whereas the method presented by Ryu \textit{et al.} \cite{confocc} considers navigation using only a 2D LiDAR sensor.

\subsection{Anomaly Detection using Machine Learning}
In addition to sensor fusion and occlusion modeling, AD using machine learning is directly related to our work. One frequently used approach to AD with machine learning involves analyzing the reconstruction error of autoencoders trained on non-anomalous data. For example, Malhotra \textit{et al.} \cite{encdecad} employs an encoder-decoder architecture that learns to reconstruct non-anomalous time-series and flags samples having high reconstruction error as anomalies. An and Cho \cite{reconsvae} leverage the variational autoencoder (VAE) to detect anomalies using a more theoretically-principled reconstruction probability instead of the reconstruction error of a generic autoencoder. Lin \textit{et al.} \cite{vaelstmhybrid} combine elements seen in other works \cite{encdecad, reconsvae}, proposing a VAE-LSTM model for time-series AD that leverages a VAE to generate features for short time windows and an LSTM to capture longer-term correlations relevant to AD. Other machine learning techniques for AD have also been studied, such as recent works \cite{csi,maskclr} that utilize contrastive learning as a method of AD by detecting out-of-distribution samples.

Much research has also investigated applying AD specifically to robotics and autonomous systems. Wyk  \textit{et al.} \cite{rtsensauto} describe a method for AD of sensors in autonomous vehicles by combining a convolutional neural network (CNN) with an adaptive Kalman filter that is paired with a failure detector. He \textit{et al.} \cite{sensredund} present an approach that detects anomalies in autonomous vehicles by exploiting redundancy among heterogeneous sensors to detect anomalies in sensor readings. In robotics, Yoo \textit{et al.} \cite{slip} describe a multi-modal autoencoder for AD applied to object slippage, whereas Park \textit{et al.} \cite{vaelstm} use an LSTM-based VAE to detect anomalies in robot-assisted feeding. In the agricultural field, prior work \cite{svae} introduced a supervised VAE-based approach operating on LiDAR and proprioperceptive measurements to predict a variety of anomalies. However, these approaches focus on predicting anomalies only as or after they have already occurred and cannot be directly used for forecasting future anomalies.

\subsection{Proactive Anomaly Detection in Robotics}
Several works have proposed proactive AD methods, which--in contrast to reactive AD methods--can enable prediction of anomalies before they occur in order to take corrective actions to avoid failure entirely or to reduce the damage caused by such a failure. LaND \cite{land} and BADGR \cite{badgr} utilize a CNN architecture operating on input images from a robot's camera. The features extracted from the CNN are used as an initial state for an LSTM which predicts future events using planned control actions as input. Most similar to our proposed method are multi-modal proactive anomaly detection methods, such as PAAD \cite{paad} and GrASPE \cite{graspe}. PAAD fuses camera images, 2D LiDAR data, and a predicted future trajectory using an attention-based multi-modal fusion architecture. GrASPE predicts navigation success probabilities for future trajectories using a multi-modal fusion architecture (fusing 3D LiDAR, RGB camera, and odometry data). The fusion mechanism in GrASPE uses graph neural networks (GNNs), forming a graph with sensor features as nodes. Sensor reliability information in GrASPE is also provided via the graph adjacency matrix (with sensor reliability computed through hand-designed, non-learning-based algorithms). However, both GrASPE and PAAD do not explicitly capture prior sensor state, with PAAD using only the sensor data from the current time step to make predictions and GrASPE relying only on a history of velocity measurements to capture the prior state of the robot.

\section{METHOD}\label{sec:method}
Our goal is to design a method to predict future failures of an autonomous field robot during operation that is robust even in cases of brief total sensor occlusion.

The model we propose accepts multi-modal sensory inputs from two sensors: a 2D LiDAR unit and an RGB camera. The 2D LiDAR produces a vector of range measurements ${\mathbf{x}_{l}^{(t)} \in \mathbb{R}^{L}}$, and the RGB camera produces images ${\mathbf{x}_{c}^{(t)} \in \mathbb{R}^{H \times W \times 3}}$. The model predicts future probabilities of failure for the next $T$ time steps based on knowledge of future controls generated by a predictive controller used by the robot. As a result, the model also requires inputs specifying such planned control actions. Following the approach of PAAD \cite{paad}, we provide the planned control actions as a grayscale image ${\mathbf{x}_{p}^{(t)} \in \mathbb{R}^{H \times W \times 1}}$, in which the planned path from the predictive controller is projected from the camera's point of view as a curve onto a blank image. To incorporate historical information that aids in prediction during periods of total sensor occlusion, our proposed model leverages a latent representation of state ${\mathbf{h}^{(t)} \in \mathbb{R}^{D}}$ that is used as input and evolved in each prediction step as new sensory and control inputs are provided. At each prediction step, the network outputs $T$ probabilities of future failure  $\mathbf{\hat{y}}^{(t:t+T)} \coloneqq (y^{(t)}, y^{(t+1)}, ..., y^{(t+T-1)}) \in [0, 1]^{T}$. For each time step, the model also predicts the probability of occlusion for the LiDAR and camera inputs: $y_{\text{lidar}}^{(t)} \in [0, 1]$ and $y_{\text{camera}}^{(t)} \in [0, 1]$.

Similar to existing works \cite{badgr, land, graspe, paad}, the proactive nature of our proposed model is beneficial by allowing prediction of future failures. In addition, like PAAD \cite{paad} and GrASPE \cite{graspe}, our model uses a variety of sensor modalities to provide improved prediction robustness. Our proposed model also explicitly models sensor occlusion. However, as compared with GrASPE, our mechanism for occlusion prediction is directly learned within the neural network model, while GrASPE uses classical (non-learning-based) algorithms to determine sensor reliability. Learning occlusion via neural network grants greater flexibility by enabling the model to learn more nuanced representations of occlusion (such as those produced by intermediate layers of an occlusion prediction network) and does not require a hand-crafted algorithm for detecting occlusion.

Unlike the prior models \cite{land, badgr, paad, graspe}, our proposed inclusion of a latent representation of robot state (which summarizes prior control and sensory inputs) allows our model to show increased resilience to false positives in cases of brief total sensor occlusion. For example, if the robot traverses a corn row with no obstructing ground-based obstacles and briefly experiences occlusion of both LiDAR and camera from leaves in the crop canopy, prior models may raise an anomaly, whereas our proposed model can utilize knowledge of the lack of obstacles captured by the latent state representation to avoid falsely reporting failures. We also combine the learning-based sensor occlusion estimation in the attention mechanism that fuses the sensory inputs, control inputs, and latent representation of state. The inclusion of occlusion estimation in the attention mechanism enables our model to learn how to combine information about predicted sensor occlusion with the latent robot state to provide improved predictions of future failure during sensor occlusion.

\subsection{Data}
We utilize the dataset collected in a prior work \cite{paad} to verify our model. This data was collected using the 4-wheeled, skid-steer TerraSentia mobile robot. The TerraSentia features a forward-facing RGB camera (OV2710) producing images with a resolution of $240 \times 320$, and a LiDAR (Hokuyo UST-10LX) with $270^{\circ}$ range at an angular resolution of $0.25^{\circ}$ that yields 1081 range measurements. The predictive path is generated using the robot's predictive controller, and is projected onto a front-facing plane using the camera's known intrinsic parameters.

In addition to the failure labels provided in the dataset, we add labels specifying camera occlusion and LiDAR occlusion. LiDAR occlusion was automatically labeled with samples displaying a median range measurement of less than $0.3$ $\text{m}$ for the center $215^{\circ}$ of LiDAR measurements being labeled as occluded. Images were automatically labeled as occluded using thresholds on image sharpness and variance of pixel values. These image occlusion labels were then inspected and refined. Such refinement ensured correct occlusion labels even when conditions like high levels of glare from the sun led the automated labeling to predict the camera as occluded (even though the path ahead could still be seen).

\subsection{Model Architecture}
\begin{figure*}[tp]
  \centering
  \includegraphics[width=7in]{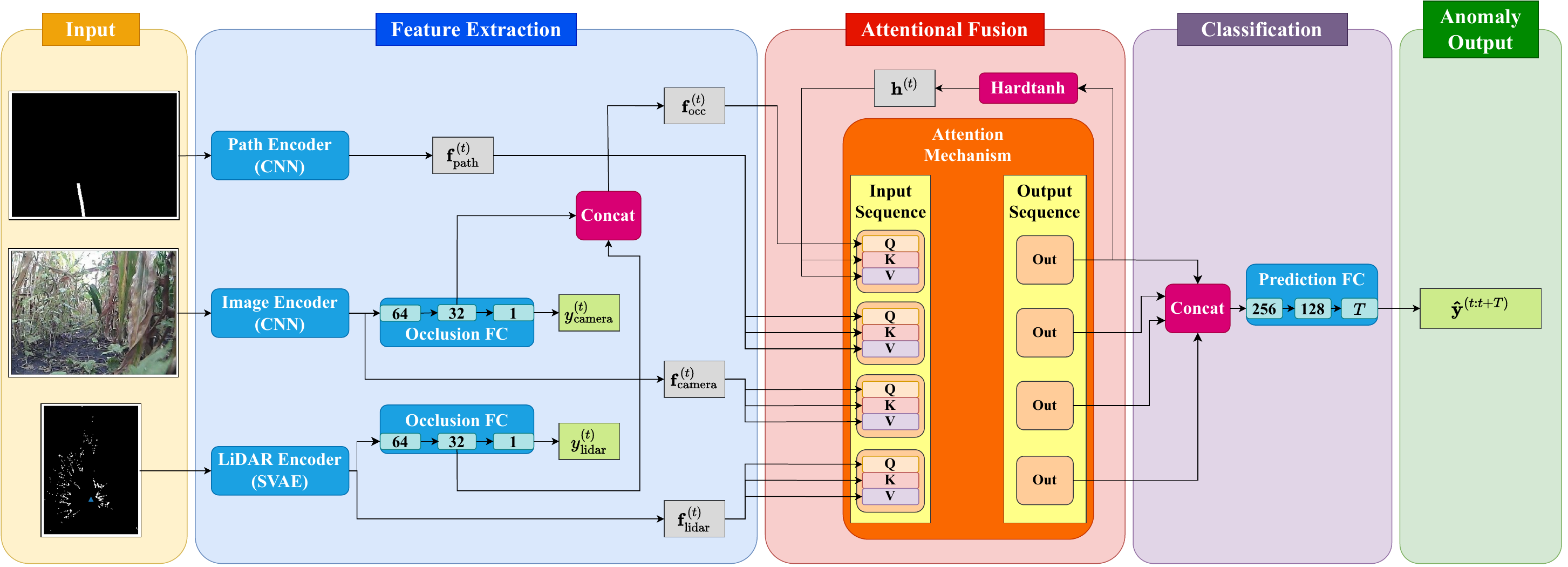}
  \caption{Proposed network architecture for ROAR. Neural networks are shown as blue boxes, intermediate features as gray boxes, non-learned operations as pink boxes, the attention mechanism as an orange box, and outputs as green boxes. For clarity and conciseness, the SVAE \cite{svae} decoder is not shown.}
  \label{networkarch}
\end{figure*}

The architecture for our model (shown in Fig. \ref{networkarch}) consists of three feature extractors, a multi-head attention fusion module, a recurrent state feature, a fully-connected occlusion prediction head for each sensory input, and a fully-connected proactive anomaly detection prediction head.

The three feature extractors are adopted from PAAD \cite{paad} as they have been shown to perform well in the agricultural field environment. The planned trajectory feature extractor accepts a $240 \times 320$ grayscale image as input and applies a region-of-interest (ROI) pooling layer followed by a convolutional neural network to produce a 64-dimensional feature vector ${\mathbf{f}_{\text{path}}^{(t)} \in \mathbb{R}^{64}}$. The RGB image feature extractor is a convolutional neural network based on a ResNet-18~\cite{resnet} backbone, with the convolutional layers pretrained on a visual navigation task \cite{visnavtask}. This image feature extractor accepts a $240 \times 320$ RGB image and outputs a feature vector ${\mathbf{f}_{\text{camera}}^{(t)} \in \mathbb{R}^{64}}$. Finally, LiDAR features are extracted with a supervised variational autoencoder (SVAE) as in prior works \cite{svae, paad}; this LiDAR feature extractor uses 1081-dimensional LiDAR input measurements to produce an output feature vector ${\mathbf{f}_{\text{lidar}}^{(t)} \in \mathbb{R}^{64}}$, where the features are the concatenated means and log-variances produced by the VAE. %variational autoencoder.

The features from the LiDAR and camera feature extractors are provided as inputs to occlusion prediction head networks, which feature two layers (the first having 32 outputs with ReLU activation and the second having 1 output with sigmoid activation). These prediction heads can be viewed as functions $g_{\text{lidar,occ}} : \mathbf{f}_{\text{lidar}}^{(t)} \mapsto y_{\text{lidar}}^{(t)}$ and $g_{\text{camera,occ}} : \mathbf{f}_{\text{camera}}^{(t)} \mapsto y_{\text{camera}}^{(t)}$.

The data from the feature extractors is fused using a multi-head attention module. The multi-head attention mechanism can be viewed as computing attention for elements in a sequence, where the sequence elements are the state features, camera features, LiDAR features, and trajectory features. The multi-head attention module utilizes 8 attention heads. The keys and values for the attention module are formed by concatenating the state vector with the features computed by the feature extractors: 
\begin{equation} \label{eq1}
K=V=[\mathbf{h}^{(t)}, \mathbf{f}_{\text{path}}^{(t)}, \mathbf{f}_{\text{camera}}^{(t)}, \mathbf{f}_{\text{lidar}}^{(t)}]
\end{equation}
For the queries, the final three elements are the same as for the keys and values. The query for the first element (corresponding to the latent state representation) is formed by concatenating occlusion-biased features from outputs of the first fully connected layer of $g_{\text{camera,occ}}$ and the first fully connected layer of $g_{\text{lidar,occ}}$, which are denoted $\mathbf{o}_{\text{camera}}^{(t)} \in \mathbb{R}^{32}$ and $\mathbf{o}_{\text{lidar}}^{(t)} \in \mathbb{R}^{32}$, respectively. Letting $\mathbf{f}_{\text{occ}}^{(t)} = [\mathbf{o}_{\text{camera}}^{(t)}, \: \mathbf{o}_{\text{lidar}}^{(t)}]$, the queries are thus given by:
\begin{equation} \label{eq2}
Q=[ \mathbf{f}_{\text{occ}}^{(t)}, \mathbf{f}_{\text{path}}^{(t)}, \mathbf{f}_{\text{camera}}^{(t)}, \mathbf{f}_{\text{lidar}}^{(t)} ]
\end{equation}
The use of occlusion-biased features for the attention query vector corresponding to the latent state is informed by the assumption that the prior history of the robot is largely irrelevant when sensors are not occluded, since anomalies can be detected using the non-occluded sensor measurements. As a result, the latent state is primarily important when sensors are occluded. Thus, we query the latent state using sensory features that are biased towards features relevant in predicting occlusion to incorporate this assumption into our model.

The recurrent latent state for the network, $\mathbf{h}^{(t)}$, is initialized to a zero vector at the beginning of an inference sequence. The value of this latent state is evolved by setting it equal to the first attention output (the first sequence element in the attention mechanism corresponds to the latent state). This construction allows the latent state for the next time step to incorporate information from the sensory inputs, planned control actions, and current latent state. In addition, at each time step, a hardtanh activation (with minimum and maximum limits of -10 and 10, respectively) is applied to the new latent state in order to prevent uncontrolled growth in the magnitude of the latent state for longer sequences.

The final prediction of the future anomaly scores is done by concatenating the features output from the attention module and feeding them through 2 fully connected layers. The first of the fully connected layers has a 128-dimensional output, upon which ReLU activation and dropout \cite{dropout} are applied (with a dropout probability of 0.5). The second fully-connected layer outputs a $T$-dimensional vector with a sigmoid activation function applied; the outputs of this fully connected layer are the final anomaly prediction probabilities for the next $T$ time steps.

\subsection{Model Training}
The model is trained with a loss function that is composed of four components, corresponding to the SVAE \cite{svae} feature extractor loss ($\mathcal{L}_{\text{SVAE}}$, which is composed of a KL divergence term and a reconstruction term), the anomaly classification loss ($\mathcal{L}_{\text{anomaly}}$), the camera occlusion classification loss ($\mathcal{L}_{\text{camera,occ}}$), and the LiDAR occlusion classification loss ($\mathcal{L}_{\text{lidar,occ}}$). The total loss is given by:
\begin{equation} \label{eq3}
\mathcal{L} = \mathcal{L}_{\text{SVAE}} + \alpha\mathcal{L}_{\text{anomaly}} + \beta\mathcal{L}_{\text{camera,occ}} + \gamma\mathcal{L}_{\text{lidar,occ}}
\end{equation}
where $\alpha$, $\beta$, and $\gamma$ are coefficients that specify the relative weighting of the individual loss terms. Based on prior work~\cite{paad}, we use a value of $\alpha = 6.21$. As we are primarily interested in the anomaly detection output, we set $\beta = 0.1\alpha$ and $\gamma = 0.1\alpha$ to prevent the training from focusing too heavily on occlusion outputs at the expense of the true output of interest.

\section{EXPERIMENTAL RESULTS}\label{sec:expresults}
Our experiments involve a modified version of the dataset from PAAD, where we add labels indicating occlusion of the LiDAR and camera, as described in Section \ref{sec:method}. This dataset features 4.1 km of navigation with the TerraSentia, where the robot moves with a reference speed of 0.6 m/s and data is logged at 3 Hz. For the experiments, the network predicts failures for the next 10 time steps (i.e., $T=10$).

We utilize the same training and test split as in PAAD, which features 29284 training samples (2262 of which involve anomalies) and 6869 test samples (of which 696 involve anomalies). Due to the small number of anomalies, we re-balance the training dataset by under-sampling non-anomalous samples and over-sampling anomalous samples. To train the sequential aspect of our model, we split the training dataset into contiguous sequences of length 8 to help with batching. The initial latent state is set to zero for the first prediction step, with the remaining 7 steps using the latent state from the prior prediction. At test time, the sequences are not split into these fixed-length sequences; instead, prediction at test time operates on entire temporally-coherent sequences, with the first prediction step using a latent state initialized to a zero vector. The model is trained using the Adam optimizer \cite{adam} with a learning rate of 0.0005 and a weight decay coefficient of 0.00015.

\subsection{Baselines}
We compare the accuracy of ROAR against the following baselines on the test set:
\begin{itemize}
    \item \textit{CNN-LSTM} \cite{land, badgr}: a model for intervention and future event prediction proposed in LaND and BADGR. This network features a convolutional feature extractor that generates features from the input RGB image, and then uses the image features as an initial hidden state in an action-conditioned LSTM that predicts the probability of future failure.
    
    \item \textit{NMFNet} \cite{nmfnet}: an anomaly detection adaptation of a multi-modal fusion network that was devised for robot navigation in difficult environments. Like in prior work \cite{paad}, we maintain the branch operating on 2D laser data and the branch operating on image data, and replace the 3D point cloud branch with a fully-connected network that processes future actions from the predictive controller.
    
    \item \textit{PAAD} \cite{paad}: the proactive anomaly detection network featured in our prior work. This network features an SVAE LiDAR feature extractor \cite{svae}, a path feature extractor CNN, a ResNet-based camera image feature extractor \cite{resnet}, a multi-head attention sensor fusion module, and a fully-connected fusion layer that combines path image features with the fused observation features.
    
    \item \textit{Graph Fusion}: a graph fusion network inspired by GrASPE \cite{graspe}, where nodes correspond to features extracted from sensor and control inputs, along with an additional state node with a self-loop that is added to inject state information (rather than using prior velocity measurements as in GrASPE). Two GCN \cite{gcn} layers (with edge weights computed as reliability measurements from the automatically-generated occlusion labels) and a GATv2 \cite {gatv2} layer are followed by a fully-connected failure prediction network.
\end{itemize}

We benchmark against Graph Fusion rather than GrASPE as GrASPE uses data sources unavailable in our dataset (e.g., 3D LiDAR), and we are particularly interested in comparing the fusion mechanism of ROAR with the graph-based fusion seen in GrASPE. To ensure fair comparison, the CNN used for computing RGB camera features in each baseline model is the same pretrained ResNet-18 used by ROAR. In addition, we show results using an image-only version of our model (\textit{IO-ROAR}) that removes the LiDAR feature extractor to highlight the ability for a variant of our network to provide results comparable to or exceeding multi-sensor alternatives even in cases where additional sensors may not
be available.

\subsection{Quantitative Results}
We evaluate the models using two quantitative metrics:
\begin{itemize}
    \item \textit{F1-score}: the harmonic mean of precision and recall, given by $F\textit{1} = 2PR/(P+R)$. This metric quantifies performance of the model in a threshold-dependent manner, where we select a threshold of 0.5 (i.e., we flag failures as when the predicted probability of failure exceeds 0.5). The F1-score varies from 0 to 1, with higher values being better.
    \item \textit{PR-AUC}: a metric calculating the area under the precision-recall curve. This is a threshold-independent metric that quantifies anomaly prediction performance.\footnote{Due to the highly skewed nature of the dataset, PR-AUC is used instead of ROC-AUC \cite{rocprauc}.} This metric varies from 0 to 1, with higher values being better.
\end{itemize}
The results are presented in Table I. To account for different initializations potentially yielding better results, the values shown in Table I are the averages for each model over 5 training runs with different random seeds. We also present the results of PR-AUC and F1-score for the best model of the 5 trainings, where the best model is selected as the model having the highest value for the threshold-independent PR-AUC metric.

\begin{table}[bp]
\caption{Performance of Different Anomaly Detection Methods}
\begin{center}
\begin{tabular}{@{} l l *{4}{r} @{}}
\toprule
Model & Modality & \multicolumn{2}{c}{Average} & \multicolumn{2}{c}{Best Model} \\
\cmidrule(lr){3-4} \cmidrule(lr){5-6}
& & \multicolumn{1}{c}{PR-AUC} & \multicolumn{1}{c}{F1} & \multicolumn{1}{c}{PR-AUC} & \multicolumn{1}{c}{F1} \\
\midrule

CNN-LSTM & Image-Only & \multicolumn{1}{c}{0.568} & \multicolumn{1}{c}{0.429} & \multicolumn{1}{c}{0.577} & \multicolumn{1}{c}{0.397} \\
%IO-PAAD & Image-Only & \multicolumn{1}{c}{0.704} & \multicolumn{1}{c}{0.519} & \multicolumn{1}{c}{0.756} & \multicolumn{1}{c}{0.540} \\
IO-ROAR & Image-Only & \multicolumn{1}{c}{0.758} & \multicolumn{1}{c}{0.574} & \multicolumn{1}{c}{0.797} & \multicolumn{1}{c}{0.599} \\

\midrule

NMFNet & Multi-Modal & \multicolumn{1}{c}{0.655} & \multicolumn{1}{c}{0.559} & \multicolumn{1}{c}{0.717} & \multicolumn{1}{c}{0.567} \\
Graph Fusion & Multi-Modal & \multicolumn{1}{c}{0.775} & \multicolumn{1}{c}{0.618} & \multicolumn{1}{c}{0.807} & \multicolumn{1}{c}{0.621} \\
PAAD & Multi-Modal & \multicolumn{1}{c}{0.790} & \multicolumn{1}{c}{0.604} & \multicolumn{1}{c}{0.817} & \multicolumn{1}{c}{0.606} \\
ROAR & Multi-Modal & \multicolumn{1}{c}{\textbf{0.800}} & \multicolumn{1}{c}{\textbf{0.639}} & \multicolumn{1}{c}{\textbf{0.834}} & \multicolumn{1}{c}{\textbf{0.692}} \\
\bottomrule
\end{tabular}
\end{center}
\end{table}

As seen in Table I, our method (ROAR) outperforms all baselines in terms of both F1-score and PR-AUC. The superior performance of ROAR compared to the baselines demonstrates how incorporating prior robot state and learned occlusion information leads to improved anomaly detection performance. Furthermore, we see that even the image-only variant of our network outperforms one of the multi-modal baselines that leverages an additional sensor (NMFNet), and performs only slightly worse than Graph Fusion and PAAD (which both leverage an additional sensor modality).

Table II shows the number of parameters and neural network inference time of Graph Fusion, PAAD, and ROAR (collected on a machine with an i9-9900K and an RTX 2070 Super). The results in Table II demonstrate that ROAR has comparable inference speed to PAAD, despite providing improved anomaly detection accuracy. ROAR also displays faster inference than Graph Fusion due to the additional overhead introduced by using graph neural networks. Graph Fusion also requires occlusion labels to be provided as inputs even at test time (which adds additional latency beyond the neural network inference time shown in Table II by requiring computation of occlusion label inputs), while ROAR does not need the occlusion labels to be provided during test time.

\begin{table}[tp]
\caption{Neural Network Inference Time and Number of Parameters of Graph Fusion, PAAD, and ROAR}
\begin{center}
\begin{tabular}{@{} l l *{3}{r} @{}}
\toprule
Model & \multicolumn{1}{c}{Network Inference Time (ms)} & \multicolumn{1}{c}{\# Parameters (million)} \\
\midrule
Graph Fusion & \multicolumn{1}{c}{5.98} & \multicolumn{1}{c}{11.98} \\
PAAD & \multicolumn{1}{c}{4.99} & \multicolumn{1}{c}{11.91} \\
ROAR & \multicolumn{1}{c}{4.99} & \multicolumn{1}{c}{11.92} \\
\bottomrule
\end{tabular}
\end{center}
\end{table}

\subsection{Qualitative Results for Total Sensor Occlusion} 
\begin{figure}[tp]
  \centering
  \includegraphics[width=3.in]{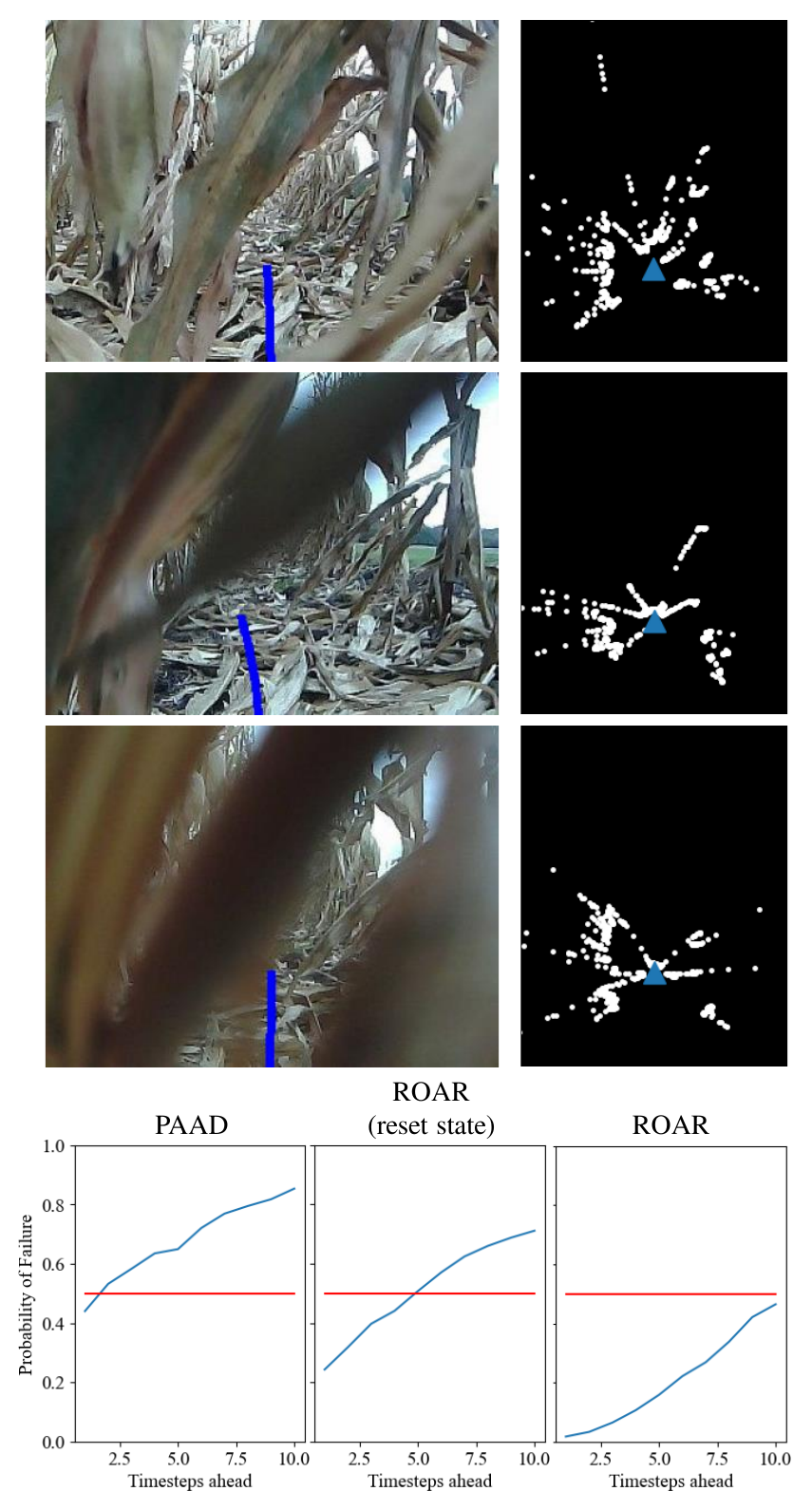}
  \caption{Image (with predicted path drawn as a blue curve) and LiDAR readings for three sequential frames, along with predictions for the last frame using PAAD, reset-state ROAR, and ROAR.}
  \label{qualitative1}
\end{figure}

In Fig. \ref{qualitative1}, we demonstrate qualitative predictions of our model in the case of total sensor occlusion. The prediction probabilities are shown as a blue curve in the graphs and an anomaly probability threshold of 0.5 is shown as a red line. The predictions shown in Fig. \ref{qualitative1} demonstrate the robustness of our model to producing false positives in cases of brief sensor occlusion in an otherwise obstacle-free environment. In Fig. \ref{qualitative1}, we also show predictions using PAAD and reset-state ROAR (a variant of ROAR in which we set the state vector to zero to highlight the effect of removing the state information). While both PAAD and reset-state ROAR produce false positives in this case, ROAR does not produce a false positive in this example occlusion scenario.

\begin{figure}[t]
  \centering
  \includegraphics[width=3.5in]{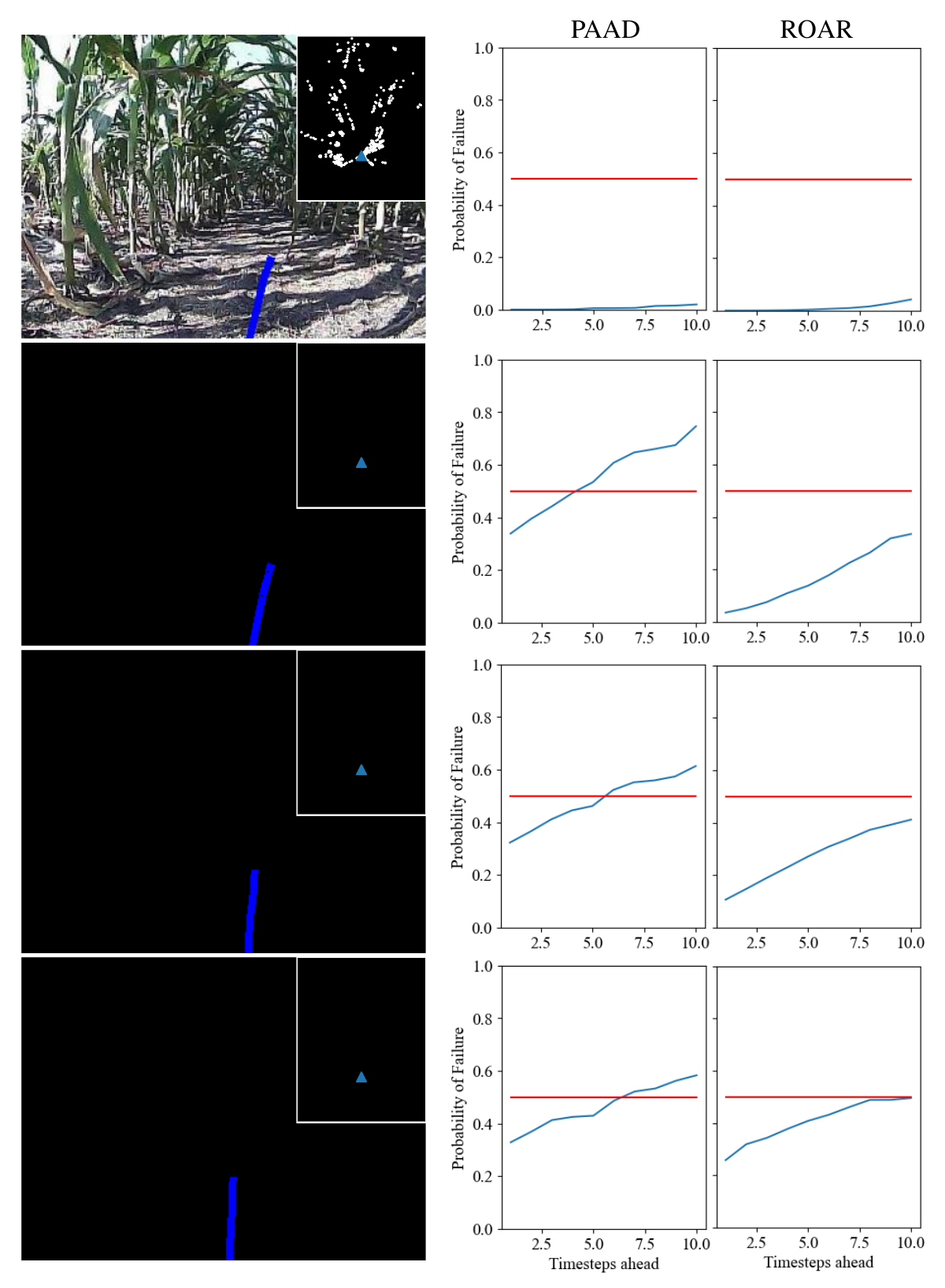}
  \caption{Predictions using PAAD (left) and ROAR (right) on a sequence featuring synthetic total sensor occlusion for the final three frames. ROAR shows greater robustness to simultaneous LiDAR and camera occlusion when compared to PAAD.}
  \label{qualitative2}
\end{figure}

Furthermore, in Fig. \ref{qualitative2}, we demonstrate the effect of prolonged synthetic total sensor occlusion for both PAAD and ROAR. In Fig. \ref{qualitative2}, the occlusion of all sensors causes PAAD to predict failures in the near future (even though the path was clear), whereas ROAR is robust to the occlusion of all sensors. However, as the period of occlusion lengthens, ROAR becomes increasingly likely to predict a failure. This demonstrates how our model captures the intuitive insight that brief occlusions in otherwise normal scenarios are not necessarily failures, but as the duration of occlusion grows, the probability of failure increases.

\subsection{Ablation Study}
To study how different design considerations affect our model, we conducted an ablation study. We specifically analyze four variants of our model:
\begin{itemize}
    \item \textit{No State}: a version of our model where the latent state is removed (always set to zero), preventing the model from capturing information about the history of sensor and control inputs.
    \item \textit{No Occlusion}: a version of our model where the occlusion modeling is removed, and the query vector for the attention module equals the key and value vectors.
    \item \textit{Fixed Occlusion}: a version of our model where we provide a vector of repeated occlusion labels produced by the automated labeling algorithm as the state query vector instead of using the occlusion-biased features (i.e., this variant includes occlusion predictions but the predictions are not learned using a neural network).
    \item \textit{ROAR}: the complete ROAR model.
\end{itemize}

\begin{table}[bp]
\caption{Ablation Study}
\begin{center}
\begin{tabular}{@{} l *{4}{r} @{}}
\toprule
Model & \multicolumn{2}{c}{Average} & \multicolumn{2}{c}{Best Model} \\
\cmidrule(lr){2-3} \cmidrule(lr){4-5}
& \multicolumn{1}{c}{PR-AUC} & \multicolumn{1}{c}{F1} & \multicolumn{1}{c}{PR-AUC} & \multicolumn{1}{c}{F1} \\
\midrule
No State & \multicolumn{1}{c}{0.766} & \multicolumn{1}{c}{0.542} & \multicolumn{1}{c}{0.783} & \multicolumn{1}{c}{0.543} \\
No Occlusion & \multicolumn{1}{c}{0.762} & \multicolumn{1}{c}{0.644} & \multicolumn{1}{c}{0.832} & \multicolumn{1}{c}{0.636} \\
Fixed Occlusion & \multicolumn{1}{c}{0.777} & \multicolumn{1}{c}{\textbf{0.651}} & \multicolumn{1}{c}{0.818} & \multicolumn{1}{c}{0.653} \\
ROAR & \multicolumn{1}{c}{\textbf{0.800}} & \multicolumn{1}{c}{0.639} & \multicolumn{1}{c}{\textbf{0.834}} & \multicolumn{1}{c}{\textbf{0.692}} \\
\bottomrule
\end{tabular}
\end{center}
\end{table}

The results from the ablation study are shown in Table III. These results show averages over 5 trainings on different random seeds to account for different initializations, as well as the metrics for the best model (the model of the 5 training runs that displays the highest value of the threshold-independent PR-AUC metric).

The ablation study shows that, compared to other variants, ROAR on average displays higher performance in terms of the threshold-independent PR-AUC, and the best ROAR model outperforms the other models both on PR-AUC and F1-score. Such results demonstrate the importance of both occlusion modeling and the use of state. These results also show that learning occlusion with a neural network and using occlusion-biased features for querying state outperforms the approach of using non-learning-based algorithms to classify occlusions and providing the resulting labels as neural network inputs. Furthermore, while the final ROAR model displays slightly lower average F1-score than the no occlusion and fixed occlusion models, the higher average PR-AUC is more advantageous due to its threshold-independent nature. Specifically, PR-AUC provides a general, threshold-independent picture of anomaly detection performance, whereas the F1-score could be improved for a fixed model by tuning the detection threshold.

\section{CONCLUSION}\label{sec:conclusion}
We have presented a novel occlusion-aware recurrent neural network architecture for proactive anomaly detection in field environments that is particularly well-suited for cases when brief periods in which all sensors are occluded are possible. Our network fuses sensory input data, a planned trajectory, and a latent representation of state to predict probabilities of future failure over a given time horizon. We further enhanced our network by explicitly learning when sensors are occluded, and using this learned information to moderate the use of our latent representation of robot state. Our experimental results validate our approach by demonstrating superior quantitative performance over prior methods, while also qualitatively showing robustness to false positives during brief periods when all sensors are occluded. Although our method outperforms the baselines, it shows the limitation of requiring explicit labels of failures due to the use of supervised learning. One possible direction for future work could be to adopt a semi-supervised or unsupervised approach, such as one based on reconstruction errors.

\addtolength{\textheight}{-12cm}   % This command serves to balance the column lengths

%%%%%%%%%%%%%%%%%%%%%%%%%%%%%%%%%%%%%%%%%%%%%%%%%%%%%%%%%%%%%%%%%%%%%%%%%%%%%%%%
%%%%%%%%%%%%%%%%%%%%%%%%%%%%%%%%%%%%%%%%%%%%%%%%%%%%%%%%%%%%%%%%%%%%%%%%%%%%%%%%
%%%%%%%%%%%%%%%%%%%%%%%%%%%%%%%%%%%%%%%%%%%%%%%%%%%%%%%%%%%%%%%%%%%%%%%%%%%%%%%%
%%%%%%%%%%%%%%%%%%%%%%%%%%%%%%%%%%%%%%%%%%%%%%%%%%%%%%%%%%%%%%%%%%%%%%%%%%%%%%%%

\bibliographystyle{IEEEtran} % We choose the "plain" reference style
\bibliography{IEEEabrv, references}

\end{document}